\documentclass[lettersize,journal]{IEEEtran}
\usepackage{amsmath,amsfonts}
\usepackage{algorithmic}
\usepackage{algorithm}
\usepackage{array}
\usepackage{caption}  % 添加caption包
\usepackage[caption=false]{subfig}  % 修改subfig设置
\usepackage{textcomp}
\usepackage{url}
\usepackage{verbatim}
\usepackage{graphicx}
\usepackage{cite}
\begin{document}
	
%\newpage
%\thispagestyle{empty}
%This work has been submitted to the IEEE for possible publication. Copyright may be transferred without notice, after which this version may no longer be accessible.
%\clearpage
\newpage
\thispagestyle{empty}
\twocolumn[%
\vspace*{\fill}
This work has been submitted to the IEEE for possible publication. Copyright may be transferred without notice, after which this version may no longer be accessible.
\vspace*{\fill}
]
\clearpage

\title{Imitation Learning Policy based on Multi-Step Consistent Integration Shortcut Model}

\author{Yu Fang, Xinyu Wang, Xuehe Zhang,\IEEEmembership{Member, IEEE}, Wanli Xue, Mingwei Zhang, Shengyong Chen,\IEEEmembership{Senior Member, IEEE}, Jie Zhao,\IEEEmembership{Senior Member, IEEE}
 
%\thanks{This paper was produced by the IEEE Publication Technology Group. They are in Piscataway, NJ.}
%\thanks{Manuscript received April 19, 2021; revised August 16, 2021.}
\thanks{ This work was supported in part by the National Natural Science Foundation of China under Grant 62373129, Grant 92048301, and in part by the Heilongjiang Provincial Natural Science Foundation of China Grant YQ2023F012. ( Yu Fang and Xinyu Wang contribute equally to this work, Corresponding authors: Xuehe Zhang, Jie Zhao)}
\thanks{Yu Fang, Xinyu Wang, Xuehe Zhang, Mingwei Zhang, and Jie Zhao are with the State Key Laboratory of Robotics and Systems, Harbin Institute of Technology, Harbin 150001, China. (e-mail: 20b308008@stu.hit.edu.cn,wxinyu@hit.edu.cn, zhangxuehe@hit.edu.cn,21B908040@stu.hit.edu.cn,jzhao@hit.edu.cn). 
	
Wangli Xue and Shengyong Chen are the School of Computer Science and Engineering, Tianjin University of Technology, Tianjin 300384, China.  (e-mail: xuewanli@email.tjut.edu.cn,sy@ieee.org).
	}
}

\markboth{Journal of \LaTeX\ Class Files,~Vol.~14, No.~8, August~2021}%
{Shell \MakeLowercase{\textit{et al.}}: A Sample Article Using IEEEtran.cls for IEEE Journals}


\maketitle

\begin{abstract}
The wide application of flow-matching methods has greatly promoted the development of robot imitation learning. However, these methods all face the problem of high inference time. To address this issue, researchers have proposed distillation methods and consistency methods, but the performance of these methods still struggles to compete with that of the original diffusion models and flow-matching models. In this article, we propose a one-step shortcut method with multi-step integration for robot imitation learning. To balance the inference speed and performance, we extend the multi-step consistency loss on the basis of the shortcut model, split the one-step loss into multi-step losses, and improve the performance of one-step inference. Secondly, to solve the problem of unstable optimization of the multi-step loss and the original flow-matching loss, we propose an adaptive gradient allocation method to enhance the stability of the learning process. Finally, we evaluate the proposed method in two simulation benchmarks and five real-world environment tasks. The experimental results verify the effectiveness of the proposed algorithm. 
\end{abstract}


\section{Introduction}
%\IEEEPARstart{I}{mitation} learning (IL) is a fundamental approach for enabling robots to acquire motor skills. It infers appropriate actions from sensory inputs—including RGB images, depth maps, and proprioceptive signals. IL has been widely applied to legged locomotion, manipulation, navigation, and grasping.
\IEEEPARstart{I}{mitation} learning (IL) is a fundamental paradigm for enabling robots to acquire motor skills by mapping sensory inputs—such as RGB images, depth maps, and proprioceptive signals—to appropriate actions. It has been widely adopted in diverse domains including impedance control~\cite{9985425} ,locomotion~\cite{7284710}, grasping~\cite{10752517} and manipulation~\cite{10305294,10432985,zhang2025boosting}.

%With advances in machine learning, imitation learning (IL) has progressed from toy demonstrations to complex, real-world tasks. Recently, diffusion models have been incorporated into IL, substantially advancing performance and robustness. However, these gains come with a cost: inference is computationally expensive. This limitation is especially restrictive for on-robot learning and deployment, where compute budgets is tight. There is a pressing need for one-step policies that retain the benefits of iterative generative methods while markedly reducing computational overhead.
With advances in machine learning in Computer Vision~\cite{11036758,9543528,10852355}, Natural Language Processing~\cite{guo2025deepseek}, and more, IL has evolved from simple toy demonstrations to complex, real-world robotic tasks. Recently, diffusion models have emerged as powerful generative backbones in IL, achieving remarkable performance and robustness. However, their iterative denoising process incurs high computational cost, making real-time deployment on physical robots impractical—especially when on-device compute budgets are tight. Therefore, there is an urgent need for efficient one-step policies that preserve the advantages of diffusion-based methods while drastically reducing computational overhead.

%To this end, robotics researchers have adapted knowledge-distillation techniques from computer vision to train robotic skill policies. However, a fundamental limitation of distillation is that the student’s performance is typically bounded by the teacher, making teacher-level accuracy a practical ceiling. Recently, Shortcut~\cite{frans2024one} was proposed as an online training framework that enforces a one-step policy and introduces an expected step-size objective.
To address this issue, robotics researchers have borrowed knowledge-distillation techniques from computer vision to train efficient skill policies. Yet, a fundamental limitation persists: the student’s performance is inherently bounded by that of the teacher, forming a practical upper ceiling. Unlike classic knowledge-distillation, recently the Shortcut framework~\cite{frans2024one} reformulates policy distillation into an online training paradigm that enforces a one-step policy and introduces an expected step-size objective.
%\begin{figure}[t]
%	\centering
%	\subfloat[diffusion policy\label{fig:sub1}]{
%		\includegraphics[width=0.4\textwidth]{diffusion_policy_1.jpg}
%	}\\  % 换行符控制纵向排列
%	\vspace{0.8em}  % 添加垂直间距（可选）
%	\subfloat[shortcut policy\label{fig:sub2}]{
%		\includegraphics[width=0.4\textwidth]{shortcut_policy_1.jpg}
%	}\\  % 换行符控制纵向排列
%	\vspace{0.8em}  % 添加垂直间距（可选）
%	\subfloat[multi consist policy\label{fig:sub3}]{
%		\includegraphics[width=0.4\textwidth]{multi_consis_policy_1.jpg}
%	}
%	% \caption{Overview of the differences between diffusion models, ShortCut models, and our method. The diffusion strategy in a) generates motion strategies through iterative denoising in small steps. The ShortCut in b), by introducing expected step lengths, can achieve denoising in fewer steps to generate motion strategies. Our proposed method in c), by averaging the one-step loss over multiple steps, increases the generation speed while ensuring the quality of the generated results.}
%	\caption{Overview of the differences among diffusion models, Shortcut models, and our approach. }
%	\label{fig:main}
%\end{figure}
\begin{figure}[htpb]
	\centering
	\centerline{\includegraphics[width=1.0\columnwidth]{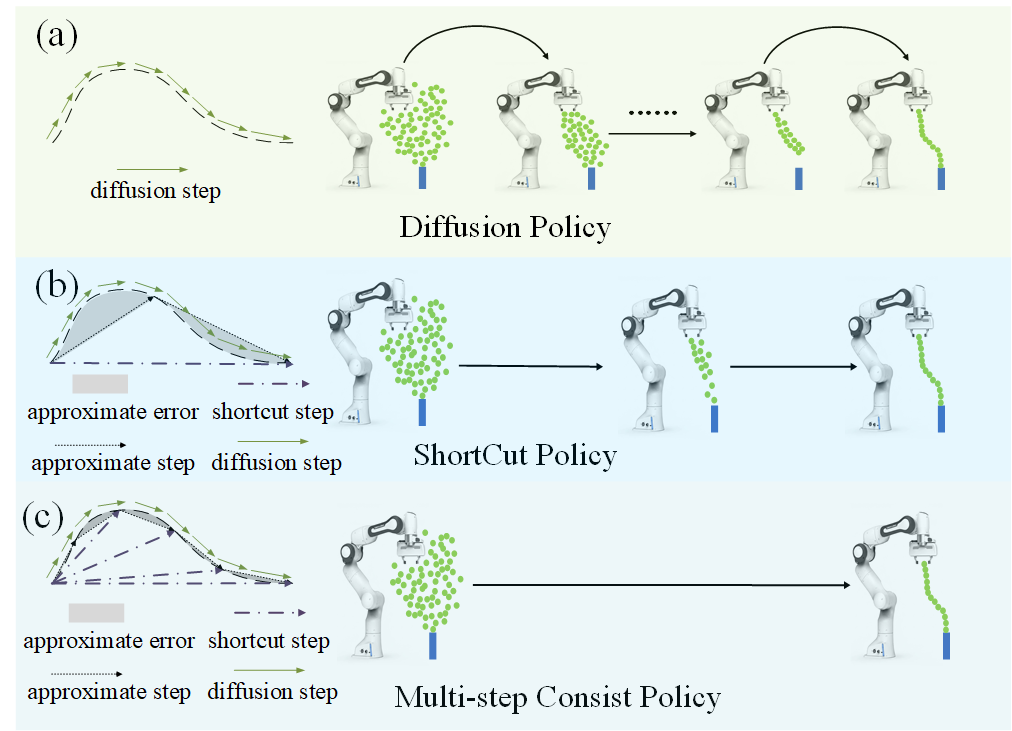}}
	\caption{Overview of the differences among diffusion models, Shortcut models, and our approach.}
	\label{fig:main}
\end{figure}

Although Shortcut demonstrates promising efficiency for robotic imitation learning, its one-step generation quality remains unsatisfactory. Shortcut can be interpreted as a hybrid between learning and distillation, where a reduced-step policy is distilled from the original flow-matching model. This coupling mode increases optimization difficulty; therefore, alleviating the learning burden becomes crucial for improving overall performance.  

To overcome this challenge, a multi-step consistency strategy is introduced to facilitate optimization by supervising the policy with multiple-step predictions in this paper. Specifically, integration is performed along the same trajectory over several short steps, decomposing the long-range shortcut into a sequence of local approximations. This design reduces cumulative approximation errors and establishes multiple constraint points along the shortcut path, effectively distributing the one-step error across multiple intervals and improving the quality of one-step generation.

Figure~\ref{fig:main} conceptually compares our method with prior approaches. As shown, diffusion models rely on iterative denoising with small steps, while Shortcut introduces an expected step length to generate motion policies with fewer iterations. Our approach further extends Shortcut by averaging one-step losses across multiple steps, improving generation speed without compromising quality. The standard Shortcut configuration can be viewed as a special case of our framework when the step number is set to two.

% 我们的方法可以看作是多步版本的shortcut，当步数设置为2就变成了标准的shortcut。
% 由于多步一致损失是一个自一致损失，它的值一般都很小，且在初始阶段几乎为0，而flow-matching loss的值比较大，从而使得多步一致损失的梯度信号很容易被掩盖。

%Since the multi-step consistency loss is a self-consistency loss, its values are generally very small and nearly zero during the initial phase, whereas the flow-matching loss tends to be significantly larger. As a result, the gradient signal of the multi-step consistency loss can easily be overshadowed. To better balance gradients from the flow-matching and consistency objectives—and prevent any single gradient from dominating the optimization—we adopt multi-task gradient-correction principles and propose a gradient-adaptive method. By reorienting the composite gradient, we reduce optimization difficulty and improve overall performance.

Because the multi-step consistency loss measures self-consistency, its magnitude is typically small—often close to zero at the beginning of training—while the flow-matching loss remains considerably larger. Consequently, the gradient signal from the consistency term may be easily dominated. To maintain a balanced contribution between the two objectives, we introduce a gradient-adaptive scheme inspired by multi-task gradient correction. This method reorients the composite gradient to mitigate imbalance, ease optimization, and improve overall performance.

%In summary, to mitigate the excessive iteration steps required by diffusion-based imitation learning, we introduce a multi-step consistency strategy. This strategy substantially improves the quality of one-step action and facilitates the deployment of diffusion models for robot policy generation. Our main contributions are summarized as follows:
In summary, to alleviate the excessive iterative steps in diffusion-based imitation learning, we propose a multi-step consistency framework that markedly improves one-step generation quality and facilitates the practical deployment of diffusion models in robot policy learning. The key contributions are as follows:

%\begin{enumerate}
%	\item We present a multi-step consistency imitation-learning algorithm that improves one-step generation quality by supervising with multi-step predictions.
%	\item We present a gradient-adaptive method to balance the flow-matching and multi-step consistency objectives, which prevents any single loss from dominating the optimization and further improves one-step inference performance.
%	\item We validate the method on robotic manipulation, demonstrating strong performance and suggesting its potential as a general-purpose generative approach.
%\end{enumerate}
\begin{enumerate}
	\item Introduce a multi-step consistency imitation-learning algorithm that enhances one-step generation by supervising with multi-step predictions.
	\item Develop a gradient-adaptive optimization scheme to balance the flow-matching and consistency objectives, preventing domination by any single loss and improving inference performance.
	\item Demonstrate the effectiveness of the proposed framework on robotic manipulation, highlighting its potential as a general-purpose imitation learning approach.
\end{enumerate}
%The remaining structure of the paper is arranged as follows: Section 2 introduces the related work relevant to this paper, Section 3 details the method of this paper, Section 4 presents the experimental part of this paper, and Section 5 summarizes the entire paper.

The remainder of this paper is organized as follows. Section 2 reviews related work, Section 3 presents the proposed method, Section 4 reports experimental results, and Section 5 concludes the paper.

\section{Related works}
\subsection{Diffusion Model For Robotics}
Numerous studies have applied diffusion models, flow matching, and related generative techniques to robot skill learning. Diffusion Policy~\cite{chi2023diffusion} significantly improves skill-learning performance on diverse manipulation tasks by introducing action chunking and other training refinements. To deploy Diffusion Policy on tasks driven by 3D point-cloud perception, researchers improve computational efficiency by simplifying point-cloud network architectures~\cite{3ddiffusion}. Furthermore, distillation-based approaches have been explored for skill learning~\cite{prasad2024consistency}.

Beyond diffusion models, the concise formulation of flow matching~\cite{lipmanflow} has motivated its adoption in robot skill learning, where it has shown strong performance~\cite{braun2024riemannian}. Unlike prior work, our method targets strong one-step inference from a single training procedure. Because training occurs once (without teacher models), performance is independent of distillation and does not rely on external teachers, avoiding teacher-limited accuracy. Moreover, our approach is orthogonal to point-cloud backbone design; improvements in point-cloud networks can be seamlessly integrated into our framework.

\subsection{One-step Inference in Diffusion Model}
To improve real-time inference with diffusion models, researchers have analyzed their theoretical properties and proposed a range of sampling-acceleration methods. Examples include DDIM~\cite{songdenoising}, EDM~\cite{karras2022elucidating}, and DPM-Solver~\cite{lu2022dpm}, as well as rectified flow matching. In parallel, knowledge distillation compresses multi-step diffusion into fewer steps—sometimes even a one-step. For the distillation objective, alternatives to L2 loss have been explored, including adversarial~\cite{sauer2024adversarial} and distribution-matching~\cite{yin2024one} objectives.

Closely related to our work is Shortcut~\cite{frans2024one}. Recently, researchers introduced an expected step length to enable end-to-end training of one-step generation, a formulation that can be viewed as self-distillation. To mitigate sampling error in Shortcut, researchers proposed higher-order Shortcut variants from an inference perspective~\cite{chen2025high}. One concurrent work with ours is Meanflow~\cite{Geng2025MeanFF}. It obtains an analytical expression for the average velocity through back-propagation of the velocity gradient. However, this method has poor stability because it relies on the Jacobian Vector Product.
% 与我们同期的一个工作是meanflow，它通过对速度的梯度反向传播获得了平均速度的解析表达式，但是该方法由于依赖JVP，其稳定性较差。

Unlike prior methods, our approach unifies training and inference within a single framework. We perform multi-step integrations along the same trajectory and average per-step integration errors across steps. This procedure yields high-quality one-step generation.
\section{Method}

Recently, many iterative generative methods—such as diffusion models and flow matching—have adopted ordinary differential equation (ODE) formulations to model the transition from a noise distribution to the data distribution. Owing to its concise formulation and the advantage of bypassing reverse-time SDE simulation, flow matching has attracted broad interest. In flow matching, \( \mathbf{x}_{t} \) is defined as the interpolation between the noise distribution \( \mathbf{x}_0 \) and the data distribution \( \mathbf{x}_1 \sim \mathcal{D}\), and \( \mathbf{v}_t=\frac{d\mathbf{x}_{t}}{dt} \) is defined as the flow velocity from the noise distribution to the data distribution. In this paper, we adopt linear interpolation to define \( \mathbf{x}_t \) and \( \mathbf{v}_t \):

\begin{equation}
	\begin{aligned}
		\mathbf{x}_t&=(1-t)\mathbf{x}_{0}+t\mathbf{x}_{1}  \\
		\mathbf{v}_t&=\mathbf{x}_1-\mathbf{x}_0
	\end{aligned}    
\end{equation}

Given noise-data pairs \( \mathbf{x}_0 \), \( \mathbf{x}_1 \), and the interpolation method, \( \mathbf{v}_t \) can be determined. However, if \( \mathbf{x}_1 \) is known, there is no need to calculate \( \mathbf{v}_t \). From the perspective of an ordinary differential equation (ODE), since \( \mathbf{v}_t = \frac{d\mathbf{x}_t}{dt} \), \( \mathbf{v}_t \) can be solved by specifying \( \mathbf{x}_t \). Thus, the flow model can be optimized by regressing the true velocity of the randomly sampled noise-data pairs \((\mathbf{x}_0, \mathbf{x}_1) \sim \mathcal{D}\):

\begin{equation}
	\begin{aligned}
		\mathcal{L}_{\mathrm{F}}(\theta)=\mathbb{E}_{\mathbf{x}_0, \mathbf{x}_1 \sim D}\left[\left\|\bar{\mathbf{v}}_\theta\left(\mathbf{x}_t, t\right)-\left(\mathbf{x}_1-\mathbf{x}_0\right)\right\|^2\right]
	\end{aligned}    
\end{equation}

After training, the learned velocity field \(\bar{\mathbf{v}}_\theta(\mathbf{x}_t, t)\) is used to evolve a sample drawn from noise \( \mathbf{x}_0 \) toward a target sample \(\mathbf{x}_1 \). Sampling proceeds via numerical integration of the ODE, i.e., iterative updates of \( \mathbf{x}_t \) with a chosen step size.

\begin{equation}
	\begin{aligned}
		\mathbf{x}_{1}&=\mathbf{x}_{0}+\int_{t=0}^{t=1}\mathbf{v}\left( \mathbf{x}_t,t\right)dt\\
		&=\mathbf{x}_{0}+\sum_{t=0}^{t=1}\mathbf{v}\left( \mathbf{x}_t,t\right)\delta t
	\end{aligned}
\end{equation}

In ODE solvers, the step size is pivotal: smaller steps usually yield better discretization accuracy but require more iterations. To mitigate this cost, Shortcut augments the flow model with an explicit step-size parameter, replacing the original output \( \mathbf{v}_t(\mathbf{x}_t,t) \) with \( \mathbf{v}_t(\mathbf{x}_t, t, d) \), where \( d\) denotes the step size. To train this parameterization, Shortcut employs a self-consistency objective:

\begin{equation}
	\begin{aligned}
		\mathbf{v}(\mathbf{x}_t,t,2d)=\frac{\mathbf{v}(\mathbf{x}_t,t,d)+\mathbf{v}(\mathbf{\hat{x}}_{t+d},t+d,d)}{2}
	\end{aligned}
\end{equation}

By incorporating this self-consistency objective into the original flow-matching objective, the entire training objective for ShortCut becomes

\begin{equation}
	\begin{aligned}
		\mathcal{L}_{\mathrm{S}}(\theta)=\mathbb{E}_{\mathbf{x}_0, \mathbf{x}_1 \sim D}&\left[\left\|\bar{\mathbf{v}}_\theta\left(\mathbf{x}_t, t,0\right)-\left(\mathbf{x}_1-\mathbf{x}_0\right)\right\|^2\right.\\
		&+\left .\|\bar{\mathbf{v}}_\theta\left(\mathbf{x}_t, t,2d\right)-\mathbf{v}(\mathbf{x}_t,t,2d)\|^2\right]
	\end{aligned}    
\end{equation}

The sampling process also changes:

\begin{equation}
	\label{eq_6}
	\begin{aligned}
		\mathbf{x}_{1}=\mathbf{x}_{0}+\sum_{t=0}^{t=1}\mathbf{v}\left( \mathbf{x}_t,d,t\right)d
	\end{aligned}    
\end{equation}

We analyze Equation~\ref{eq_6} and find that \( v_t \) is no longer the instantaneous velocity at time \( t \), but the average velocity between \((t, t+d)\). Therefore, ShortCut essentially unifies the objectives of the inference and training processes. However, since the self-consistency loss term of ShortCut is based on two small step sizes, their accuracy determines the accuracy of larger step sizes. When the step size is too large, the extrapolation error becomes too large, making it difficult for ShortCut to estimate accurately. Thus, an effective approach is to reduce the extrapolation error. Consequently, we consider a more general inference process on the same trajectory:
\begin{equation}
	\label{eq_7}
	\begin{aligned}
		\mathbf{x}_{t+nd}=&\mathbf{x}_{t}+\mathbf{v}\left(\mathbf{x}_t,d,t\right)d+\mathbf{v}\left(\mathbf{x}_{t+d},d,t+d\right)d\\
		&+\cdots+ \mathbf{v}\left(\mathbf{x}_{t+(n-1)d},d,t+(n-1)d\right)d
	\end{aligned}
\end{equation}
% 我们将这个过程拆开可以写成如下：
We can break down this process and write it as follows:
\begin{equation}
	\label{eq_8}
	\begin{aligned}
		\mathbf{\hat{x}}_{t+nd}&=\mathbf{\hat{x}}_{t+(n-1)d}+ \mathbf{v}\left(\mathbf{\hat{x}}_{t+(n-1)d},d,t+(n-1)d\right)d\\
		\mathbf{\hat{x}}_{t+(n-1)d}&=\mathbf{\hat{x}}_{t+(n-2)d}+ \mathbf{v}\left(\mathbf{\hat{x}}_{t+(n-2)d},d,t+(n-2)d\right)d\\
		&\cdots \\
		\mathbf{\hat{x}}_{t+2d}&=\mathbf{\hat{x}}_{t+d}+ \mathbf{v}\left(\mathbf{\hat{x}}_{t+d},d,t+d\right)d\\
		\mathbf{\hat{x}}_{t+d}&=\mathbf{x}_{t}+ \mathbf{v}\left(\mathbf{x}_{t},d,t\right)d\\
	\end{aligned}
\end{equation}

%观察等式~\ref{eq_8}，我们发现根据积分过程，我们可以将一段长积分约束为多段较短积分，每一段积分都可以复用之前的积分项，这样我们可以对同一个轨迹上的n个积分点计算(n-1)次约束，由于每一段积分项的约束，使得每段的外推误差都得到了约束。我们将目标写为shortcut的形式：

% Observing equation~\ref{eq_8}, we find that according to the integration process, we can decompose a long integration into multiple shorter integrations, each of which can reuse the previous integration terms. In this way, we can calculate \( (n-1) \) constraints for \( n \) integration points on the same trajectory. Due to the constraints on each integration term, the extrapolation error in each segment is constrained. We formulate the objective in the form of ShortCut:
From Equation~\ref{eq_8}, the integration over a long horizon can be decomposed into multiple shorter integrations, each reusing the previous integration terms. Consequently, for \(n\) integration points along the same trajectory, we obtain \( (n-1) \) consistency constraints. Because each segment is constrained, the per-segment extrapolation error is also constrained. We formulate the training objective in the Shortcut style:

\begin{equation}
	\begin{aligned}[b]
		nd\mathbf{v}\left(\mathbf{x}_{t},nd,t\right)=&\underbrace{d\cdot\mathbf{v}\left(\mathbf{x}_t,d,t\right)+d\mathbf{v}\left(\mathbf{x}_{t+d},d,t+d\right)}_{2d\mathbf{v}\left(\mathbf{x}_{t},2d,t\right)}+\cdots\\
		&\underbrace{\quad\quad\quad+ d\mathbf{v}\left(\mathbf{x}_{t+(n-2)d},d,t+(n-2)d\right)}_{(n-1)d\mathbf{v}\left(\mathbf{x}_{t},(n-1)d,t\right)} \\
		&+d\mathbf{v}\left(\mathbf{x}_{t+(n-1)d},d,t+(n-1)d\right)
		\\
	\end{aligned}
\end{equation}

Our loss function consists of two parts :

\begin{equation}
	\begin{aligned}
		\mathcal{L}_{\mathrm{M}}(\theta)=\{\mathbb{E}_{\mathbf{x}_0, \mathbf{x}_1 \sim D}&\underbrace{\left[\left\|\bar{\mathbf{v}}_\theta\left(\mathbf{x}_t, t,0\right)-\left(\mathbf{x}_1-\mathbf{x}_0\right)\right\|^2\right.}_{{\text{Flow Matching Loss:}\mathcal{L}_{FM}}}\\
		&,\underbrace{\sum_{k=2}^{n}\left.\|\bar{\mathbf{v}}_\theta\left(\mathbf{x}_t, t,kd\right)-\mathbf{v}(\mathbf{x}_t,t,kd)\|^2\right]}_{\text{Multi Consis Loss:}\mathcal{L}_{MC}}\}
	\end{aligned}    
\end{equation}

The training procedure of our algorithm is summarized in the Algorithm~\ref{alg1}.

\begin{algorithm}[t]
	\caption{Multi-Step Consistency Flow Matching Training}
	\begin{algorithmic}
		\WHILE{not converged}
		\STATE \(\mathbf{x}_0 \sim \mathcal{N}(0,\mathcal{I}),\mathbf{x}_1\sim D,t\sim p(t),d\sim p(d),c,k,\textbf{v\_list}\)\\
		\STATE \(\mathbf{x}_t=t \mathbf{x}_0 + (1-t)\mathbf{x}_1\)
		\FOR{first m samples}
		\STATE \(\mathbf{v}_t=\mathbf{x}_1-\mathbf{x}_0\)
		\STATE \(d\longleftarrow 0\)
		\ENDFOR
		\FOR{other samples}
		\FOR{ n in range k-1}
		\IF{n==0}
		\STATE \(\textbf{v\_list.append}( \mathbf{\hat{v}}_{\theta}(\mathbf{x}_t,t,d))\)
		\STATE \(\mathbf{\hat{x}}_{t+d}\longleftarrow \mathbf{x}_{t}+\mathbf{\hat{v}}_{\theta}(\mathbf{x}_{t},t,d)d\)
		\ELSE
		\STATE \(\textbf{v\_list.append}(\mathbf{\hat{v}}_{\theta}(\mathbf{\hat{x}}_{t+(n-1)d},t+(n-1)d,d))\)
		\STATE \(\mathbf{\hat{x}}_{t+nd}\longleftarrow \mathbf{\hat{x}}_{t+(n-1)d}+\mathbf{\hat{v}}_{\theta}(\mathbf{\hat{x}}_{t+(n-1)d},t+(n-1)d,d)d\)
		\ENDIF
		\ENDFOR
		\ENDFOR\\
		\FOR{n in range(2,k+1) }
		\STATE \(d\longleftarrow nd\)
		\STATE \(\mathbf{v}_t\longleftarrow \frac{1}{m}\sum_{i=0}^{m}\text{v\_list}[i]\)
		\ENDFOR 
		\STATE \(\theta\longleftarrow \nabla_{\theta}\|\mathbf{\bar{v}_{\theta}(\mathbf{x}_t,t,d)}-\mathbf{v}_t\|^2 \)
		\ENDWHILE
	\end{algorithmic}
	\label{alg1}
\end{algorithm}

In the robot imitation-learning setting, our goal is to map noise \(\mathbf{a}_0\) to a target action distribution \(\mathbf{a}_1\). Accordingly, we define a conditioned action-generation velocity field \(\mathbf{\bar{v}_{\theta}(\mathbf{a}_t,t,d,\mathbf{o})}\). Given demonstrations \(\{\mathbf{o},\mathbf{a}_1\}\) and noise samples \(\mathbf{a}_0\), our training objective comprises two components:

\begin{equation}
	\begin{aligned}
		\mathcal{L}_{\mathrm{IL}}(\theta)=\{&\mathbb{E}_{\mathbf{a}_0, \mathbf{a}_1,\mathbf{o} \sim D}\left[\left\|\bar{\mathbf{v}}_\theta\left(\mathbf{a}_t,\mathbf{o}, t,0\right)-\left(\mathbf{a}_1-\mathbf{a}_0\right)\right\|^2\right.\\
		&,\sum_{k=2}^{n}\left.\|\bar{\mathbf{v}}_\theta\left(\mathbf{a}_t,\mathbf{o}, t,kd\right)-\mathbf{v}(\mathbf{a}_t,\mathbf{o},t,kd)\|^2\right]\}
	\end{aligned}    
\end{equation}
% 在机器人模仿学习中，低维动作空间表现出比图像更弱的分布结构，使得网络更难区分噪音和动作样本。此外，因为多步一致性目标取决于流量匹配目标，所以在联合训练期间天真地混合多步一致性和流量匹配梯度会导致梯度不平衡。具体而言，多步一致性损失通常规模较小，因此流匹配损失主导优化，使多步一致性目标训练不足并降低单步性能。在我们的设置中，多步目标的内部依赖性，与更低维度的行动相结合，进一步加剧了训练的难度。
\begin{figure}[t]
	\centering
	\subfloat[block hammer beat\label{loss1}]{
		\includegraphics[width=0.4\textwidth]{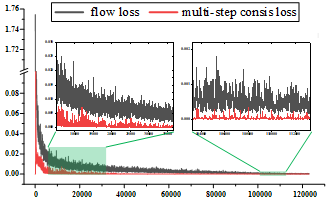}
	}
	\hfil
	\subfloat[bottle adjust\label{loss2}]{
		\includegraphics[width=0.4\textwidth]{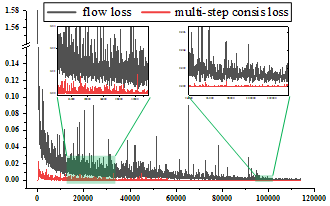}
	}
	\caption{The difference between the two losses is too large in value.}
	\label{fig:loss}
\end{figure}
In robot imitation learning, low-dimensional action spaces exhibit weaker distributional structure than images, making it harder for networks to discriminate noise from action trajectory samples. Moreover, because the multi-step consistency objective depends on the flow-matching objective, naively mixing multi-step consistency and flow-matching gradients during joint training induces gradient imbalance. Specifically, the multi-step consistency loss is typically smaller in scale, as shown in Figure~\ref{fig:loss}, so the flow-matching loss dominates optimization, leaving the multi-step consistency objective under-trained and degrading one-step performance. In our setting, internal dependencies within the multi-step objective, combined with even lower-dimensional actions, further exacerbate training difficulty.

To address this issue, we cast the integrated objective as a multi-task learning (MTL) problem. Building on established MTL methods—PCGrad~\cite{NEURIPS2020_3fe78a8a}, dynamic weight averaging (DWA)~\cite{dwa}, and IMTL~\cite{imtl}—we propose a new Adaptive Gradient Allocation (AGA) algorithm. Although the multi-step consistency and flow-matching objectives are related, their dependence is asymmetric: the multi-step consistency loss depends on the flow-matching loss, not vice versa. This asymmetry makes standard MTL schemes suboptimal for this configuration, motivating our AGA design.
\begin{figure}[t]
	\centering
	\subfloat[\label{fig2:sub1}]{
		\includegraphics[width=0.14\textwidth]{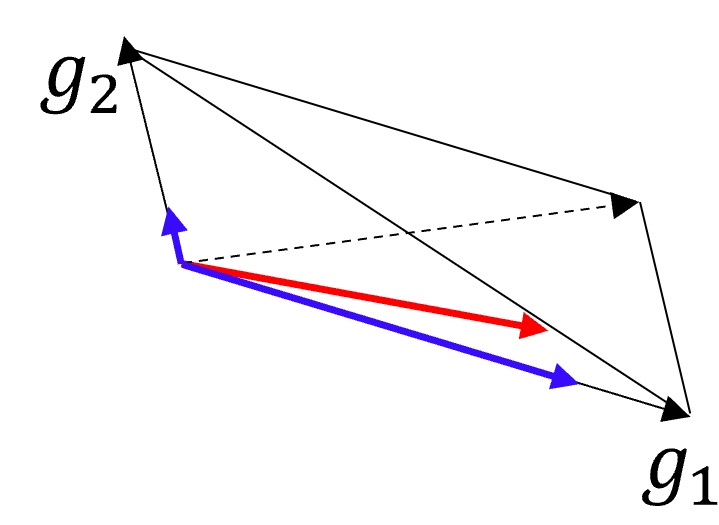}
	}
	\subfloat[\label{fig2:sub2}]{
		\includegraphics[width=0.14\textwidth]{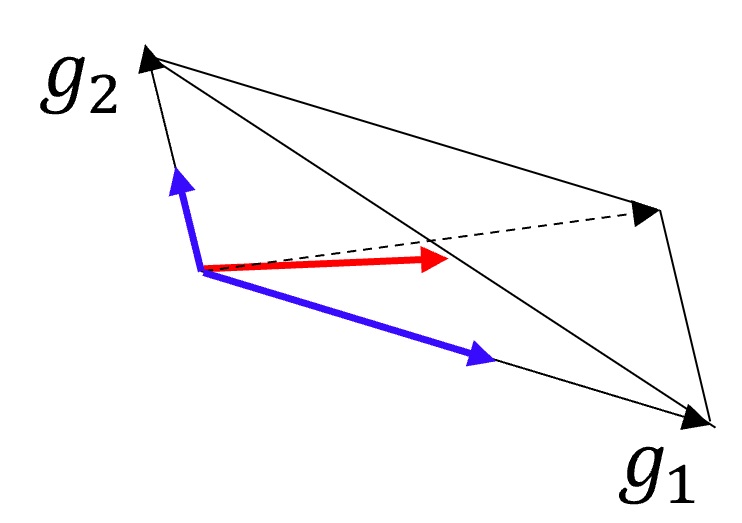}
	}
	\subfloat[\label{fig2:sub3}]{
		\includegraphics[width=0.14\textwidth]{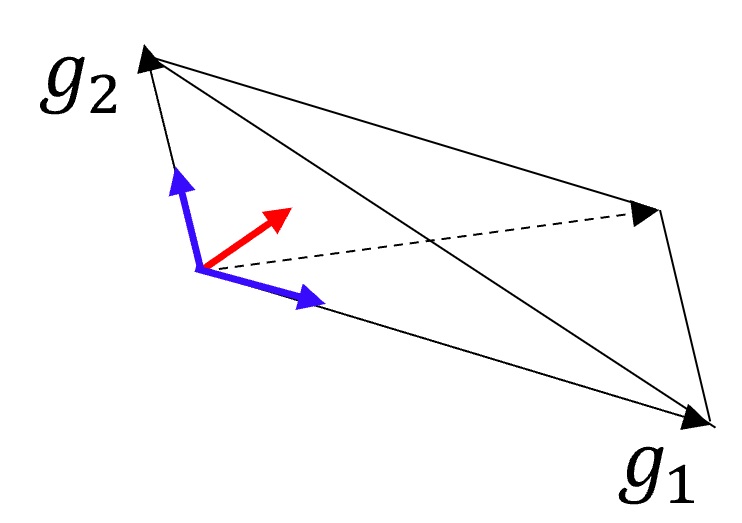}
	}
	% \caption{By adjusting the scaling factor \( c \), we can control the projection intensity of the composite gradient in different directions. This enables us to flexibly adjust the learning direction of the tasks during the gradient descent process. Figures a) to c) show that the direction of the composite gradient gradually shifts from the direction of \( \mathbf{u}_1 \) to the direction of \( \mathbf{u}_2 \). The red arrows indicate the composite gradient, while the blue arrows represent the projections of the composite gradient in different directions.}
	\caption{By adjusting the scaling factor \( c \), we can control the projection intensity of the composite gradient in different directions. Red arrows denote the composite gradient; blue arrows denote its projections onto different directions. }
	\label{fig:grad}
\end{figure}

The proposed AGA introduces an adaptive gradient-composition mechanism that adjusts the gradient direction as learning progresses across the two interdependent tasks. As shown in Fig.~\ref{fig:grad}, Figs.~\ref{fig2:sub1}–\ref{fig2:sub3} illustrate that the composite-gradient direction gradually shifts from \(\mathbf{u}_1\) toward \(\mathbf{u}_2\).  Specifically, the projections of the composite gradient onto the respective task subspaces must satisfy the following criteria:
\begin{equation}
	c(\mathbf{g}\mathbf{u}^{\top}_1)=\mathbf{g}\mathbf{u}^{\top}_2
	\label{c_eq}
\end{equation}

Let \(\mathbf{u}_1 \in \mathbb{R}^d\) denote the unit gradient direction induced by the flow-matching loss, and \(\mathbf{u}_2 \in \mathbb{R}^d\) the corresponding unit direction from the multi-step consistency loss. The composite gradient vector is formulated as \(\mathbf{g} = \alpha_1\mathbf{g}_1 + \alpha_2\mathbf{g}_2\), where \(\mathbf{g}_1 = A\mathbf{u}_1\) and \(\mathbf{g}_2 = B\mathbf{u}_2\) correspond to the gradient of the flow-matching loss and multi-step consistency loss, respectively. Here, \(A = \|\nabla_{\theta}\mathcal{L}_{\text{FM}}\|_2\) and \(B = \|\nabla_{\theta}\mathcal{L}_{\text{MC}}\|_2\) represent the \(L_2\)-norm magnitudes of their respective gradient vectors.

Then we can obtain the closed-form solution:

\begin{equation}
	\alpha_1 = \frac{B (1 - c \delta)}{A (c - \delta) + B (1 - c \delta)}, \quad \alpha_2 = 1 - \alpha_1
	\label{eq_alpha}
\end{equation}

Here, \(\delta\) is the cosine of the angle between the two gradient vectors, and the detailed derivation process is presented in Appendix A.

Based on equation \ref{c_eq}, we can specify \(c\) to adjust the strength of the composite gradient in the directions of the two tasks. To ensure that the tasks can be automatically adjusted according to the actual situation, we introduce the loss descent rate \(v_{i} = \frac{\mathcal{L}_i^{\text{current}}}{\mathcal{L}_{i}^{\text{last}}} \) for regulation.  We aim to increase the projection in the direction of the gradient generated by the multi-step consistency loss when its descent rate is relatively slower than that of the flow-matching loss (i.e.,\(v_2\) is greater, indicating a slower descent). Thus, we introduce the intensity adaptive adjustment strategy:
\begin{equation}
	c_{new}=c\exp\left(\gamma(\frac{v_2}{v_1}-1)\right)
	\label{new_c}
\end{equation}

Finally, we use Exponential Moving Average (EMA) to adjust \( c \):
\begin{equation}
	c=\beta c+(1-\beta) c_{new}
	\label{update_c}
\end{equation}

We can derive the constraint relationship between \( c \) and the gradients based on the constraints \(\alpha_1 \in (0,1)\). If the condition is satisfied, we proceed with the adaptive adjustment. If not, we simply set \(\alpha_1=0.5\) and \(\alpha_2=0.5\). The relationship that \( c \) should satisfy with the gradients is as follows:

\begin{equation}
	\left\{
	\begin{aligned}
		&\delta < c <  \frac{A\delta-B}{A-B\delta} \quad &\textbf{if} \quad \delta>0 \quad \textbf{and} A<B\delta\\
		&\max\left\{\delta,\; \frac{A}{B}\right\} \quad < c \quad &\textbf{if} \quad \delta>0 \quad \textbf{and} A=B\delta\\
		&\max\left\{\delta,\; \frac{A\delta-B}{A-B\delta}\right\} \quad < c \quad &\textbf{if} \quad \delta>0 \quad \textbf{and} A>B\delta\\
		&0 < c \quad &\textbf{if} \quad\delta \leq 0
	\end{aligned}
	\right.
	\label{delta_c}
\end{equation}

The detailed derivation process is presented in Appendix B.

In addition, we found that the multi-step consistency loss cannot provide effective guidance (usually 0) at the initial stage of the task. Therefore, in the early stage of the task, we directly employ the pcgrad~\cite{NEURIPS2020_3fe78a8a} method. Overall, the detailed gradient adaptive algorithm is presented in Algorithm\ref{alg2}.

\begin{algorithm}[t]
	\caption{Adaptive Gradient Allocation}
	\begin{algorithmic}
		\STATE \text{Input} \(\theta, \{\mathcal{L}_{FM},\mathcal{L}_{MC}\},\beta\)
		\STATE \(\mathbf{g}_{1}\longleftarrow \nabla_{\theta} \mathcal{L}_{FM}\), \(\mathbf{g}_{2}\longleftarrow \nabla_{\theta} \mathcal{L}_{MC}\)
		\FOR{ \(k\) in range \(N_{step}\)}
		\IF{\(k<n_{start}\)}
		\STATE \(\mathbf{g}^{bk}_{1}\longleftarrow \mathbf{g}_{1}\), \(\mathbf{g}^{bk}_{2}\longleftarrow \mathbf{g}_{2}\)
		\IF{\(\mathbf{g}_{2}\cdot \mathbf{g}^{bk}_{1} < 0\)}
		\STATE \(\mathbf{g}^{bk}_{1}=\mathbf{g}^{bk}_{1}-\frac{\mathbf{g}_{2}\cdot \mathbf{g}^{bk}_{1}}{\|\mathbf{g}_{2}\|^2}\cdot \mathbf{g}_{2}\)
		\ENDIF
		\IF{\(\mathbf{g}_{1}\cdot \mathbf{g}^{bk}_{2} < 0\)}
		\STATE \(\mathbf{g}^{bk}_{2}=\mathbf{g}^{bk}_{2}-\frac{\mathbf{g}_{1}\cdot \mathbf{g}^{bk}_{2}}{\|\mathbf{g}_{1}\|^2}\cdot \mathbf{g}_{1}\)
		\STATE\(\alpha_1=0.5,\alpha=0.5\)
		\STATE \(\mathbf{g}_{1}\longleftarrow \mathbf{g}^{bk}_{1}\), \(\mathbf{g}_{2}\longleftarrow \mathbf{g}^{bk}_{2}\)
		\ENDIF
		\ELSE
		\STATE compute \(c_{new}\) using Equation~\ref{new_c}
		\STATE update \(c\) using Equation~\ref{update_c}
		\IF{ c satisfies Equation~\ref{delta_c}}
		\STATE compute \(\alpha_1\) and \(\alpha_2\) using \ref{eq_alpha}
		\ENDIF
		\ELSE
		\STATE\(\alpha_1=0.5,\alpha=0.5\)
		\ENDIF
		\RETURN \(\Delta \theta=\alpha_1\mathbf{g}_1+\alpha_2\mathbf{g}_{2}\)
		\ENDFOR
	\end{algorithmic}
	\label{alg2}
\end{algorithm}

\begin{figure}[htpb]
	\centering
	\centerline{\includegraphics[width=0.85\columnwidth]{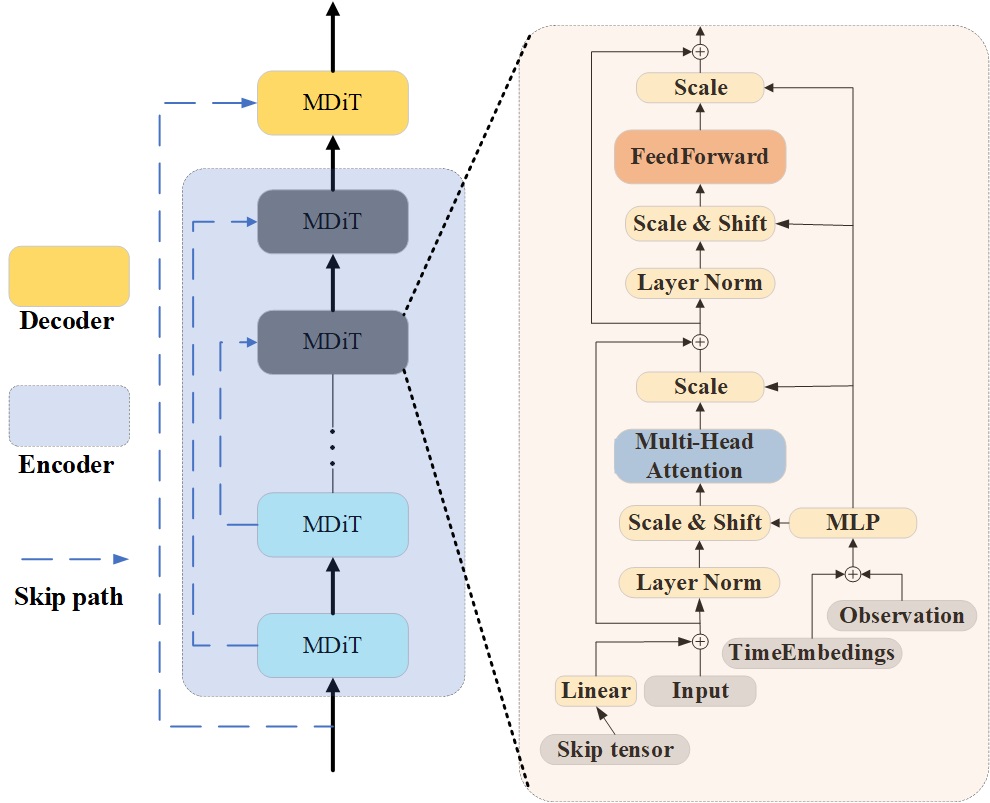}}
	\caption{UDiT1d architecture.}
	\label{udit1d}
\end{figure}

Finally, the velocity field is solved using the previously proposed UDiT1d architecture in the~\cite{11090155}. The network architecture is shown in Figure~\ref{udit1d}. Unlike our earlier setup in the~\cite{11090155}, this study does not employ self-supervised learning. Accordingly, the sliding decoder is removed, leaving a standard encoder–decoder architecture. All other components remain unchanged, including the Rotary Positional Encoding and addLLN.

During inference, occasional draws of low-probability samples can degrade performance. To mitigate this, inspired by the codebook in~\cite{DDCM}, we construct a prior codebook sampled from a Gaussian distribution. For both the multi-step consistency target and one-step inference, sampling is restricted to the codebook, thereby avoiding low-probability draws.
\section{Experiment}
\begin{figure*}[h]
	\centering
	\subfloat[block hammer beat]{{\includegraphics[width=1.4in]{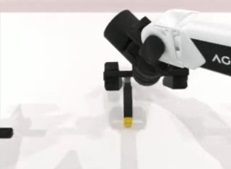}}} \hfil
	\subfloat[bottle adjust]{{\includegraphics[width=1.4in]{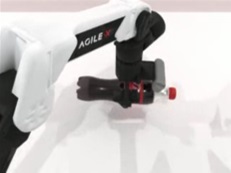}}}  \hfil
	\subfloat[container place]{{\includegraphics[width=1.4in]{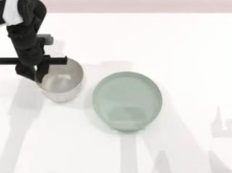}}}  \hfil
	\subfloat[diverse bottles pick]{{\includegraphics[width=1.4in]{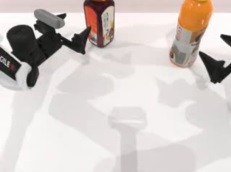}}}  \hfil
	\vfil
	\subfloat[dual bottles pick]{{\includegraphics[width=1.4in]{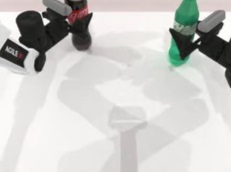}}} \hfil
	\subfloat[dual shoes place]{{\includegraphics[width=1.4in]{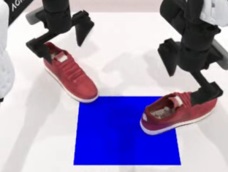}}}  \hfil
	\subfloat[mug hanging]{{\includegraphics[width=1.4in]{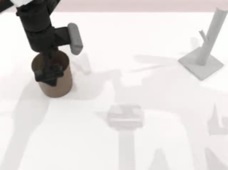}}}  \hfil
	\subfloat[pick apple messy]{{\includegraphics[width=1.4in]{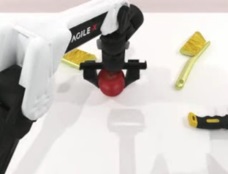}}}  \hfil
	\caption{The experiments in the RoboTwin benchmark.}
	\label{figrobowtin}
\end{figure*}
%This section evaluates the proposed method across simulated and real-world tasks. In simulation, two benchmarks with point-cloud and state observations assess performance on standard and long-horizon manipulation. We use Diffusion, Shortcut, and Meanflow~\cite{Geng2025MeanFF} as baselines. Similarly to the simulation environment, in the real world, we also evaluate multiple tasks to assess the performance of the proposed method on ordinary manipulation tasks and long-horizon manipulation tasks. In addition, in the real world, we evaluate the effectiveness of the method proposed in this paper on tasks with severe external disturbances. To probe high-frequency contact behavior, we additionally evaluate peg-in-hole and wiping tasks augmented with tactile sensing.
This section evaluates the proposed method across simulated and real-world tasks. In simulation, two benchmarks with point-cloud and state observations assess performance on standard and long-horizon manipulation. Similarly to the simulation environment, in the real world, we also evaluate multiple tasks to assess the performance of the proposed method on ordinary manipulation tasks and long-horizon manipulation tasks. In addition, in the real world, we evaluate the effectiveness of the method proposed in this paper on tasks with severe external disturbances. To probe high-frequency contact behavior, we additionally evaluate peg-in-hole and wiping tasks augmented with tactile sensing.
% 本节评估了模拟任务和真实任务中提出的方法。在模拟中，两个具有点云和状态观察的基准评估了标准和长期操作的性能，我们使用Diffusion、Shortcut以及Meanflow作为baseline。类似于仿真环境，在真实世界中，我们也评估多个任务，以评估所提出的方法在普通操作任务和长视野操作任务上的性能。此外，在真实世界中，我们在具有严重外部干扰的任务上测试了本文提出的方法的有效性。为了探索高频接触行为，我们还评估了用触觉感知增强的钉孔和擦拭任务。在真实世界中，我们使用Diffusion和Shortcut作为我们的baseline。

In summary, this section primarily addresses the following questions:

\begin{enumerate}
	\item Can the proposed method solve general manipulation tasks and long-horizon manipulation tasks in the simulation environment?
	\item How does the proposed method perform in real-world tasks?
	% \item Despite improving inference speed, how does the proposed method handle external changes?
	\item  Can the proposed method be generalized to multimodal fusion tasks?
	\item Can the handling of gradients truly enhance the performance of the algorithm?
\end{enumerate}
\subsection{Simulation Environments}

\subsubsection{RoboTwin~\cite{robotwin}}{ is a robot benchmark suite specifically designed for bimanual robot tool usage and human-robot interaction scenarios, generating diverse demonstration data through traditional planners, as shown in~\ref{figrobowtin}. We collected 50 demonstration trajectories for each task, using the point cloud as the observation.}

\subsubsection{Franka Kitchen~\cite{gupta2020relay}}{ focuses on testing long sequence tasks. It uses states as observation information, rather than point clouds or images. It contains 7 interactive objects and a manual demonstration data set with 566 demonstrations, and each demonstration completes 4 tasks in any order, as shown in~\ref{figfrankakitch}. The goal is to perform as many demonstration tasks as possible, regardless of the order, and display both short-term and long-term multimodal. }

\begin{figure}[htpb]
	\centering
	\subfloat[]{{\includegraphics[width=1.3in]{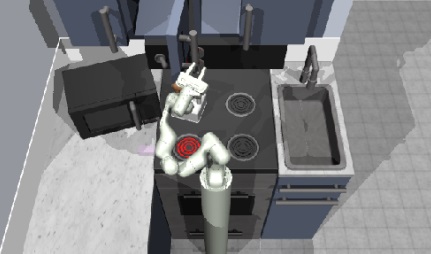}}} \hfil
	\subfloat[]{{\includegraphics[width=1.3in]{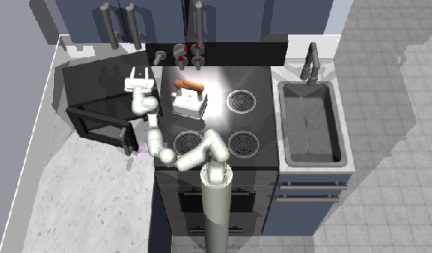}}}
	\caption{The experiments in the Franka Kitchen. a) The robot interacts with the kettle, burner, slider, and cabinet. b) The robot interacts with the kettle, microwave, top burner, and lights. }
	\label{figfrankakitch}
\end{figure}

We selected Diffusion, Meanflow and Shortcut as baselines. For Diffusion, we used the UNet architecture as the noise prediction network, while others including Shortcut,Meanflow and our proposed method utilized the UDiT1d as the velocity prediction network. It should be noted that in Robotwin, we employed the 3D-diffusion policy (3DP), whereas in Franka Kitchen, we used the Diffusion Policy (DP). In Robotwin, following the evaluation protocol of Robotwin, we collected 50 successful demonstrations. Using 3D point clouds as observations, each algorithm was trained three times with three different random seeds for 3000 epochs. The last epoch's checkpoint was selected for evaluation, which was conducted 100 times to calculate the success rate. In the Franka Kitchen environment, we used the dataset provided by Diffusion Policy. Each algorithm was trained for 3000 epochs with three different random seeds. The last epoch's checkpoint was chosen for evaluation, which was performed 100 times to calculate the success rate of interactions with different objects.

\begin{table}
	\begin{center}
		\caption{Scores on simulated task with the point cloud observation on RoboTwin. We report an average of 300 episodes and a standard deviation of 3 seeds.}
		\label{table1}
		\setlength{\tabcolsep}{3pt}
		\begin{tabular}{p{84pt}p{38pt}p{38pt}p{38pt}p{35pt}}
			\hline
			Task&3DP&Meanflow&ShortCut&Ours\\
			\hline
			block hammer beat&64.7$\pm$10.1&24$\pm$12.8&44.7$\pm$1.7& \textbf{81.3$\pm$0.9}\\
			bottle adjust&\textbf{71.7}$\pm$13.8&62.3$\pm$9.7&59.3$\pm$6.1&59.0$\pm$15.7\\
			diverse bottles pick& 32.3$\pm$10.1&51.0$\pm$5.7&35.7$\pm$4.2&\textbf{55.0$\pm$1.4}\\
			dual bottles pick(easy)&74.7$\pm$2.9&86.6$\pm$4.9&45.0$\pm$9.7&\textbf{89.3$\pm$0.5}\\
			dual bottles pick(hard)&\textbf{48.0}$\pm$7.9&48$\pm$5.7&39.7$\pm$5.4&46.7$\pm$1.7\\
			dual shoe place &7.7$\pm$2.1&11$\pm$4.3&9.4$\pm$4.1&\textbf{12.7$\pm$3.3}\\
			pick apple messy&12.7$\pm$5.5&12.6$\pm$4.2&13.3$\pm$7.6&\textbf{13.7$\pm$9.2}\\
			container place&\textbf{77.7$\pm$2.5} &65.6$\pm$4.5&65.6$\pm$2.5&55.0$\pm$5.7\\
			mug hanging(easy)&\textbf{14.0$\pm$2.3}&7.3$\pm$4.6&2.6$\pm$7.6&13.3$\pm$4.1\\
			mug hanging(hard)&10.7$\pm$3.1 &7$\pm$5.7&1.7$\pm$1.2&\textbf{16.7$\pm$1.7}\\
			\hline
			NFE & 10 &1&1 &1\\
			\hline
		\end{tabular}
	\end{center}
\end{table}
\begin{table}
	\begin{center}
		\caption{Scores on Franka Kitchen with the state observation. We report an average of 300 episodes and a standard deviation of 3 seeds.}
		\label{table2}
		\setlength{\tabcolsep}{3pt}
		\begin{tabular}{p{45pt}p{45pt}p{45pt}p{45pt}p{45pt}}
			\hline
			Task&DP&Meanflow&ShortCut&Ours\\
			\hline
			p1&\textbf{100$\pm$0.0}&\textbf{100$\pm$0.0}&99.3$\pm$0.5&\textbf{100$\pm$0.0}\\
			p2&\textbf{100$\pm$0.0}&99$\pm$0.0&99.3$\pm$0.05&\textbf{100$\pm$0.0}\\
			p3&99.7$\pm$0.5&98$\pm$0.8&98.3$\pm$1.2&\textbf{100$\pm$0.0}\\
			p4&\textbf{98.7$\pm$0.5}&94$\pm$0.8&96.3$\pm$1.7&98.0$\pm$1.4\\
			\hline
			NFE & 100 &1&1 &1\\
			\hline
		\end{tabular}
	\end{center}
\end{table}

The experimental results on the Robotwin benchmark are shown in Table~\ref{table1}. Our method outperform the one-step Shortcut in all 10 tasks and Meanflow in 7 task, even surpass the 3DP with 10 iterations in 6 tasks. With an average success rate of 44.27\% across the 10 tasks, our proposed method outperformed the 3DP's 41.39\%, which fully demonstrates the strong competitiveness of the algorithm proposed in this paper. The experimental results on the Franka Kitchen benchmark are shown in Table~\ref{table2}. In terms of object interaction, our method outperform the one-step Shortcut and Meanflow, it is comparable to the DP with 100 iterations, showing strong competitiveness. In summary, the experimental results in simulation indicate that the proposed method is highly competitive in various task scenarios.

\subsection{Real-world Environments}
\begin{figure*}[t]
	\centering
	\subfloat[ Object classification]{
		\includegraphics[width=0.32\textwidth]{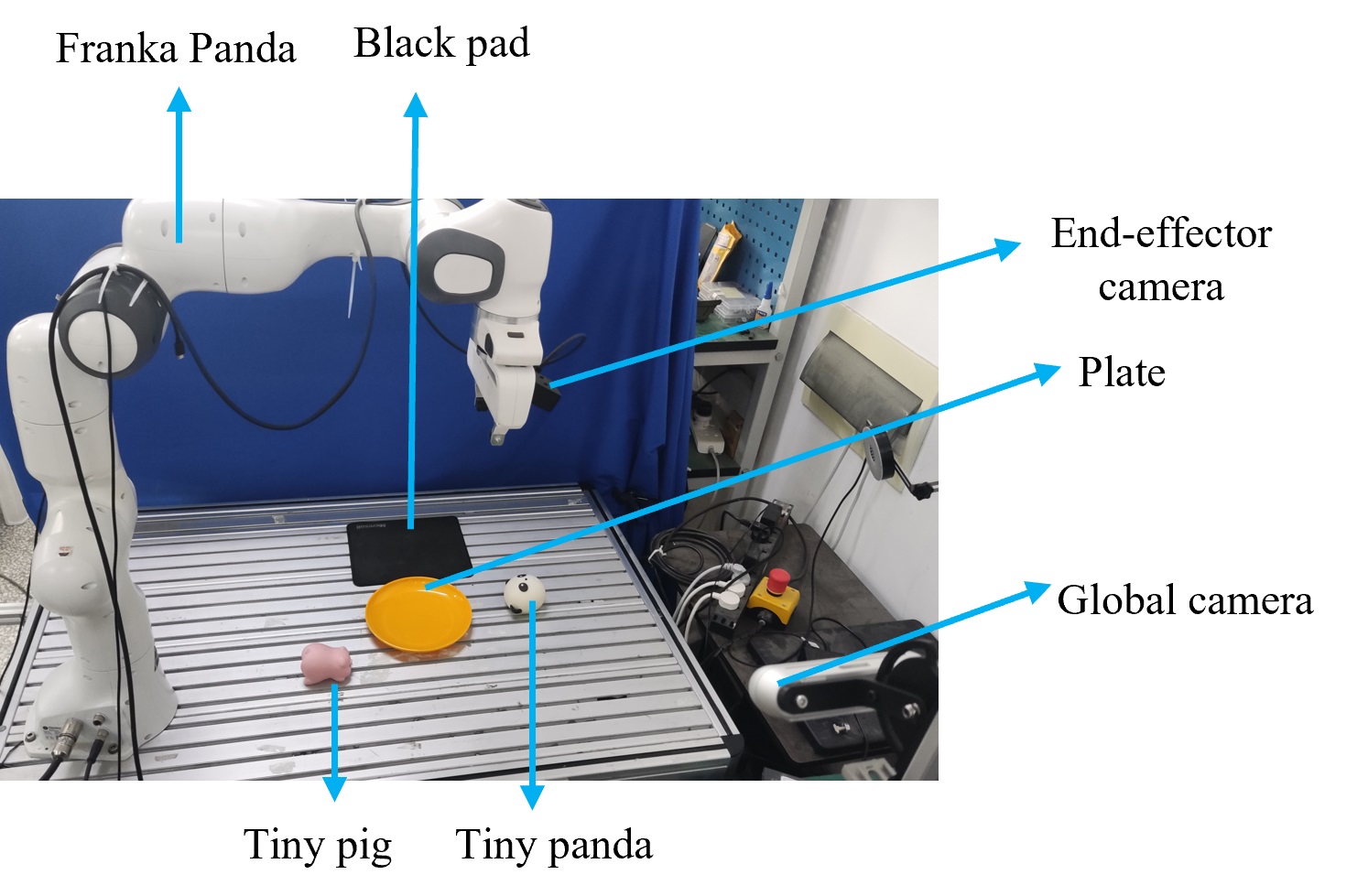}
	}
	\subfloat[ Meal packaging]{
		\includegraphics[width=0.32\textwidth]{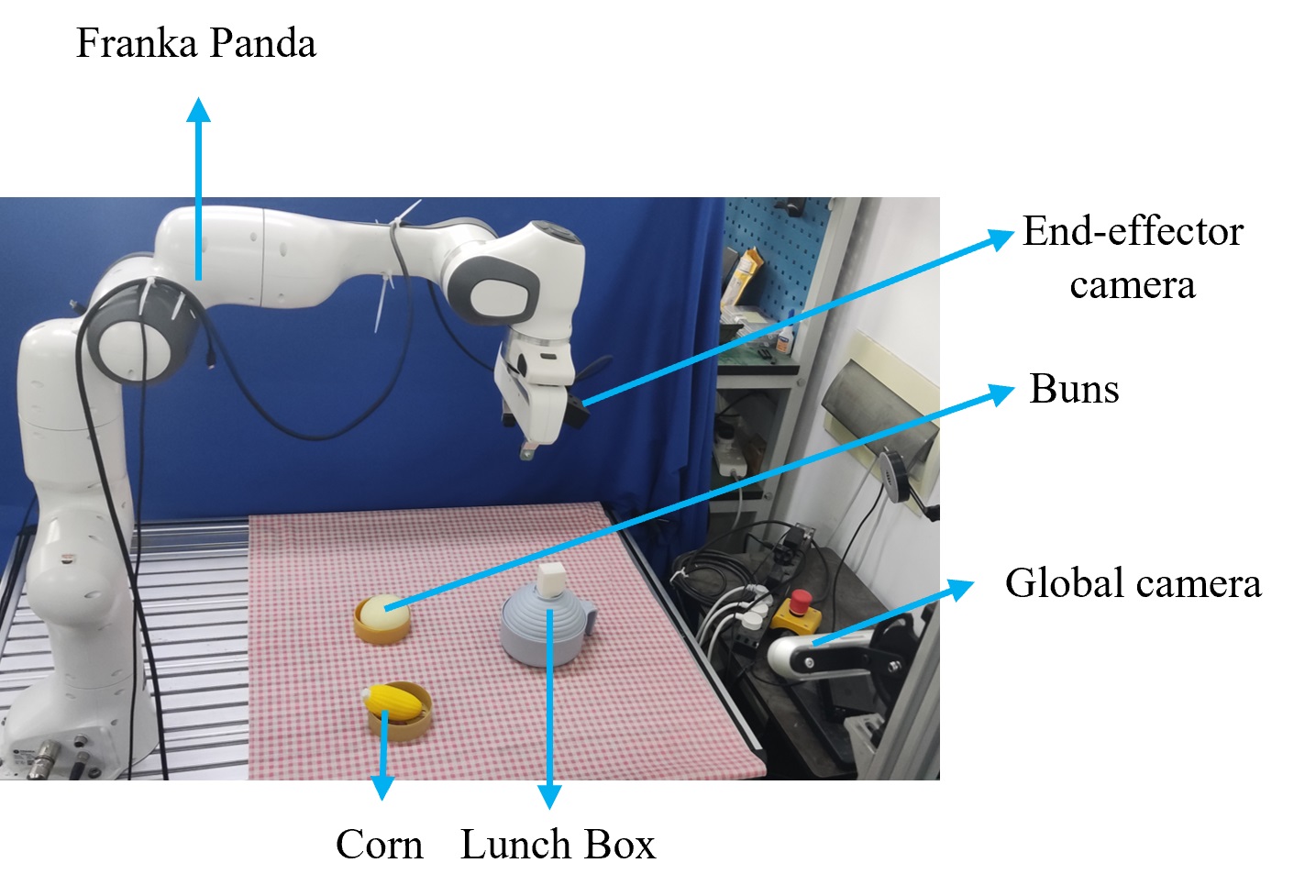}
	}
	\subfloat[ Dynamic placement]{
		\includegraphics[width=0.32\textwidth]{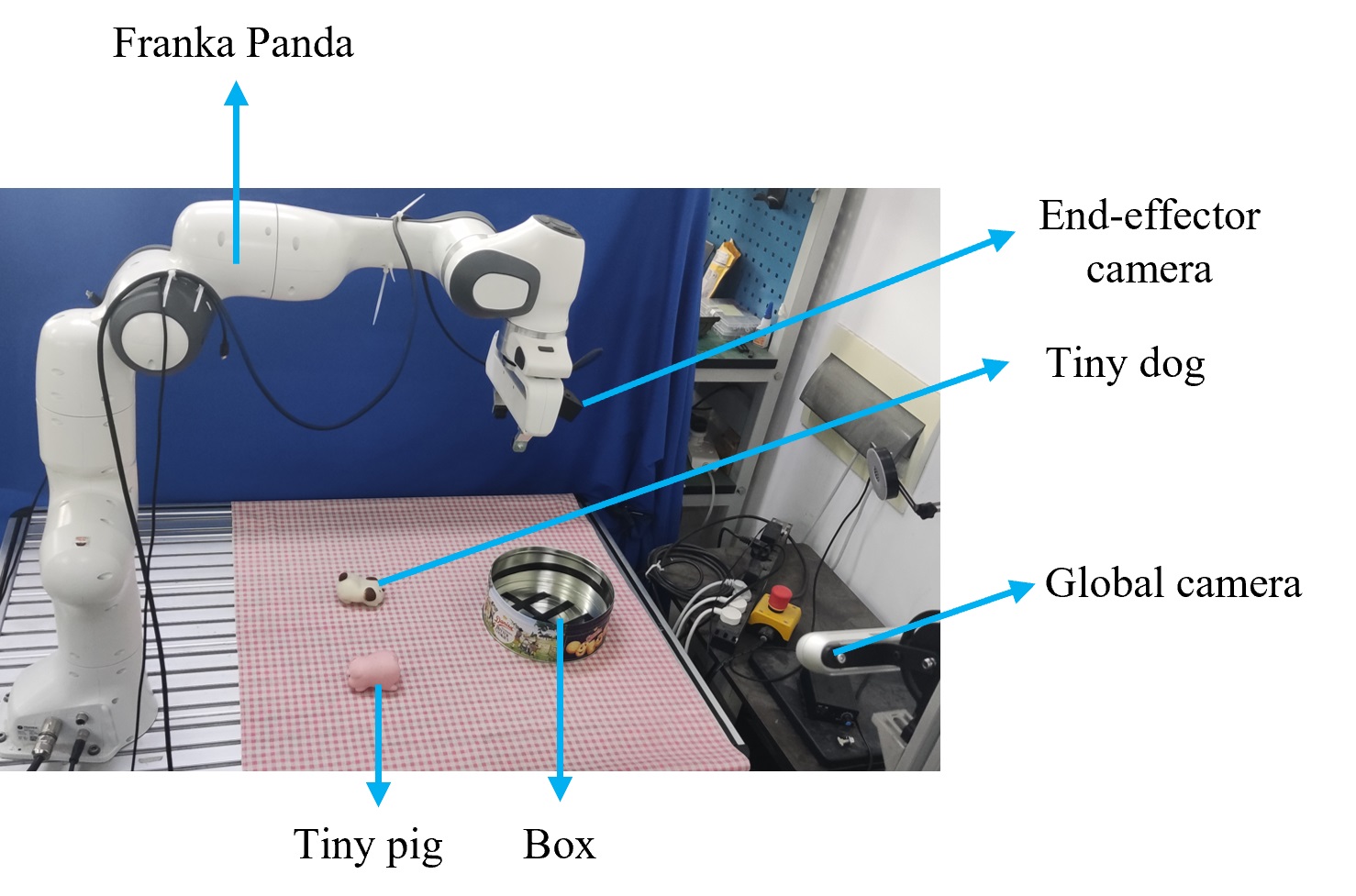}
	}
	\caption{Evaluation in the real world.}
	\label{fig:realvision}
\end{figure*}

We evaluated the effectiveness of the proposed method in this paper on tasks that utilize images as input for perception. We use the DP and the Shortcut as baselines. We selected three tasks: object classification and meal packaging, and dynamic placement tasks.

\subsubsection{Object classification}{In this task, the robot needs to learn to place different randomly placed objects into different designated areas, which is a very conventional test task. We used two cameras, the D455 and D405, for image observation. We employed Gello as the data acquisition tool and collected 81 complete and successful sets of experimental data, with a data acquisition frequency of 20 Hz.}
%在这个任务中，机器人需要学会将随机放置的不同的物体放置到不同的指定区域，这是一个非常常规的测试任务。我们使用了D455和D405两个相机作为图像观测，我们使用Gello作为数据采集工具，收集了81组完整且成功的实验数据，数据采集频率为20Hz。
\subsubsection{Meal packaging}{ In this task, the robot needs to first open the lunch box and then place the corn and buns into it, with the positions of the lunch box, corn, and buns all being random. This is a long-horizon task similar to Franka Kitchen, and the operations have a specific sequence, as the lunch box must be opened first before any subsequent actions can be performed. Like in object classification, we used two cameras, the D455 and D405, for image observation and collected 92 complete and successful demonstration datasets, with a data acquisition frequency of 20 Hz.}
% 在这个任务中，机器人需要先将饭盒打开，然后将玉米和包子放到饭盒中，其中饭盒、玉米、包子的位置都是随机的。这是一个与Franka Kitchen类似的长视界任务，并且操作具备先后顺序，因为必须首先打开饭盒，才能进行后续的操作。与object classification一样，我们使用D455和D405两个相机作为图像观测，收集了92组完整且成功的示教数据，数据采集频率为20Hz。
\subsubsection{Dynamic placement}{ In this task, the robot needs to place two different randomly placed objects into a designated box, with both the objects and the box being randomly positioned, and the box's position also being subject to human interference during task execution. This task is primarily designed to evaluate the algorithm's agility in quickly adapting to changes in the external environment. Similar to object classification, we used two cameras, the D455 and D405, for image observation, with a data acquisition frequency of 20 Hz, and collected 100 complete and successful demonstration datasets. There is no human interference in the collected demonstration datasets.}

For these three tasks, we all used the AdamW optimizer, trained for 500 epochs, and set the learning rate to 1.0e-4. We utilized the weights from the last checkpoint to test the performance of the algorithm. For each task, we conducted evaluations 10 times and then calculated the average performance. Our experimental results are shown in Table~\ref{tablereal}, which demonstrates that our method exhibits strong competitiveness across all three tasks. Specifically, our method outperforms Diffusion Policy in terms of performance and requires fewer NFE (number of function evaluations). The experimental videos can be viewed in Part I of the supplemental video.

\begin{table}
	\begin{center}
		\caption{Scores on real-world task.}
		\label{tablereal}
		\setlength{\tabcolsep}{3pt}
		\begin{tabular}{p{90pt}p{35pt}p{35pt}p{35pt}}
			\hline
			Task&DP&ShortCut&Ours\\
			\hline
			Object classification&40\%& 10\% &\textbf{50\%}\\
			Meal packaging&30\%& 10\%&\textbf{40\%}\\
			Dynamic placement&40\%&20\%&\textbf{50\%}\\
			\hline
			% NFEs & 100 &1 &1\\
			% \hline
		\end{tabular}
	\end{center}
\end{table}
\subsection{Multimodal tasks in Real-world}

% 为了进一步探究本文所提出的方法能否扩展到基于多模态的机器人操作任务中，我们结合触觉感知在插孔和擦拭两个任务上进行了实验。关于多模态感知部分，我们使用了两个resnet18处理视觉和触觉信息，然后使用一个transformer将他们融合到一起，技能学习部分，我们还是沿用UDiT1d的网络结构。任务的详细信息如下：

To further explore whether the method proposed in this paper can be extended to multimodal-based robotic manipulation tasks, we conducted experiments on two tasks: peg-in-hole and wiping, which incorporate tactile perception. For the multimodal perception component, we used two ResNet-18 models to process visual and tactile information, respectively, and then employed a transformer to fuse these two types of information. For the skill learning component, we still adopt the network structure of UDiT1d. The detailed information on these tasks is as follows:

\begin{figure}[htpb]
	\centering
	\subfloat[Peg-in-hole]{{\includegraphics[width=0.85\columnwidth]{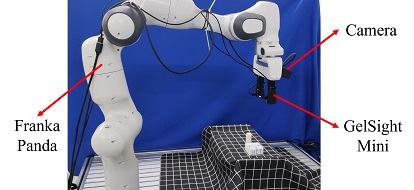}}} \vfil
	\subfloat[Wiping]{{\includegraphics[width=0.85\columnwidth]{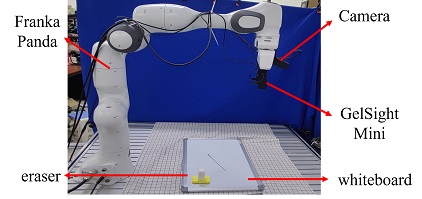}}}
	\caption{The experiments in the real world with multimodal tasks. }
	\label{figfrankakitch}
\end{figure}

\subsubsection{Peg-in-hole}{The peg-in-hole task requires picking up a cylinder and inserting it into a base accurately, which is a common visual-tactile manipulation task. For the perception part, we use a camera mounted at the end of the robotic arm to capture the manipulation scene and a GelSight Mini installed on the gripper to sense the tactile information during the contact process. We collected 80 sets of successful and complete experimental data, with a data acquisition frequency of 20 Hz.}

% 在Wiping任务中，机器人需要先抓起作业区域的板擦，然后擦除白板上的黑线。这个任务相较于peg-in-hole任务更难，因为机器人需要以合适的力与白板保持长时间的接触。与peg-in-hole任务一样，在该任务中，我们使用机械臂末端的相机获取视觉信息，使用使用安装在夹爪上的gelsightmini获取触觉信息。我们收集106组成功的完整的演示，数据采集频率为20Hz。

\subsubsection{Wiping}{In this task, the robot first needs to grasp the eraser in the working area and then erase the black lines on the whiteboard. This task is more difficult than the peg-in-hole task, as the robot is required to maintain prolonged contact with the whiteboard using an appropriate force. Similarly to the peg-in-hole task, in this task, we use a camera mounted at the end of the robotic arm to acquire visual information and a GelSight Mini installed on the gripper to obtain tactile information. We collected 106 sets of successful and complete demonstrations, with a data acquisition frequency of 20 Hz.}

\begin{figure}[htpb]
	\centering
	\centerline{\includegraphics[width=0.8\columnwidth]{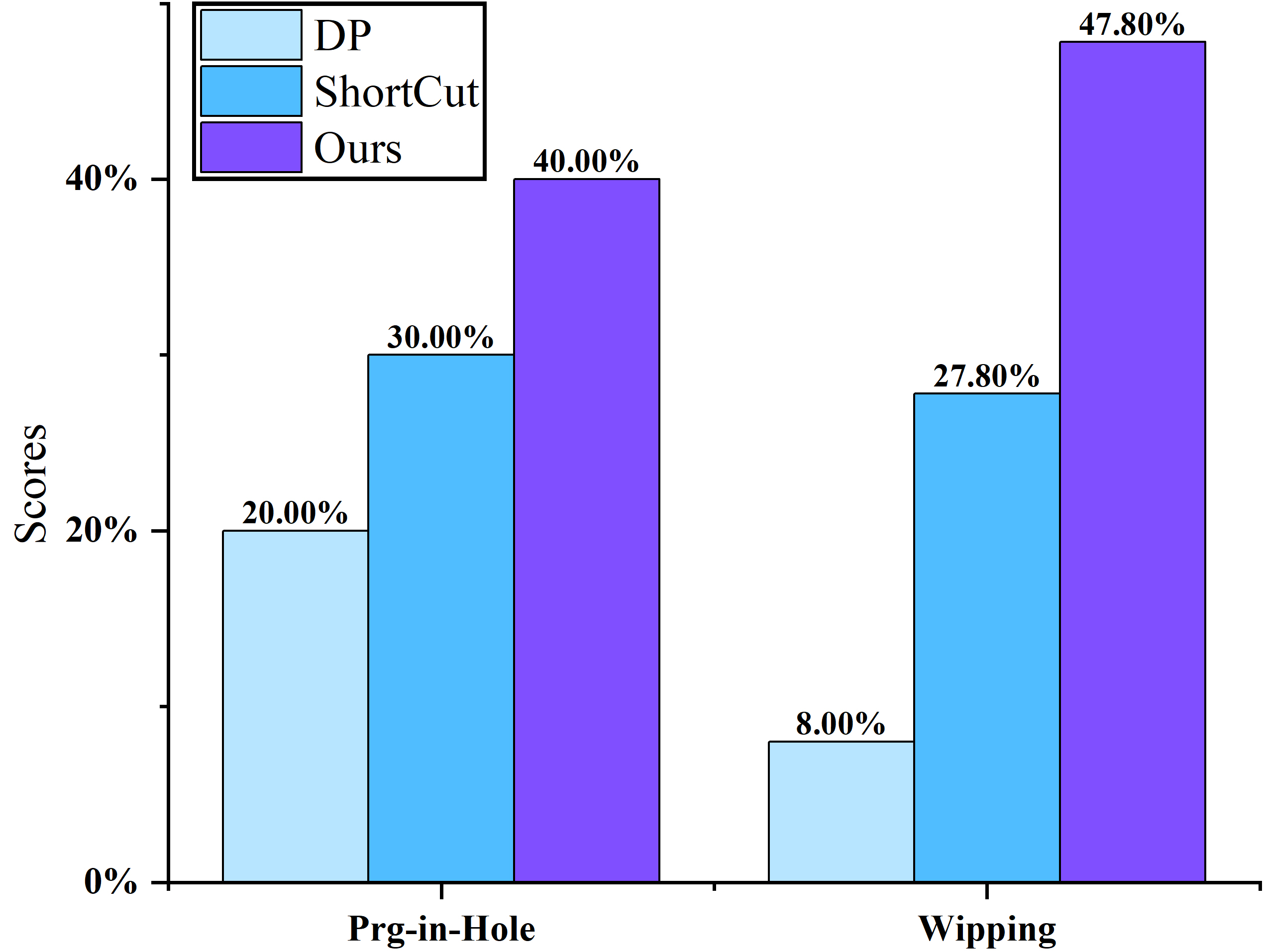}}
	\caption{Scores on real-world task with Multimodal.}
	\label{scorevt}
\end{figure}

We evaluate the performance of the algorithm on the Peg-in-Hole task using the success rate and assess its performance on the Wiping task using the completion rate, which refers to the proportion of the erased area relative to the entire black line area. For each task, we conducted 10 tests in a real-world environment and then calculated the average performance; the experimental results are shown in Fig~\ref{scorevt}. Our method achieved a 40\% success rate on the Peg-in-Hole task and a 47.8\% completion rate on the Wiping task. The algorithm’s performance on both tasks significantly outperformed that of the diffusion policy. We also found that the method using a shortcut substantially outperformed the diffusion policy. We attribute this to the algorithm’s short inference time, which enables it to demonstrate excellent performance in complex interactive tasks. The experimental videos can be viewed in Part II of the supplemental video.

\subsection{Ablation}

To further investigate the influence of computing consistency losses with different step numbers during the training phase on the final performance of the algorithm. To control the proportion of data involved in the multi-step consistency loss, as suggested in the shortcut, we select 1/4 to 1/8 of the data in each batch to calculate the multi-step consistency loss, and the remaining data is used to calculate the standard flow matching loss. We selected six groups. The first group selects 8 steps throughout the training process. The second group selects 4 steps throughout the training process. The third group uses 8 steps first and then 2 steps. The fourth group uses 4 steps first and then 2 steps. The fifth group randomly selects among 8, 4, and 2. The last group uses 4 steps first, then 8 steps, and finally 2 steps. We also conducted relevant experiments on five tasks on Robotwin. The detailed experimental results are shown in the Table~\ref{tabstep}. We found that splitting into 4 steps is relatively reasonable, and appropriately mixing the number of steps is also effective.

\begin{table*}[!t]
	\begin{center}
		\caption{The influence of computing consistency losses with different step numbers}
		\label{tabstep}
		\setlength{\tabcolsep}{3pt}
		\begin{tabular}{p{85pt}p{42pt}p{42pt}p{42pt}p{42pt}p{42pt}p{42pt}}
			\hline
			Task &8&4&8-2&4-2&random&4-8-2\\
			\hline
			bottle adjust&40.0\(\pm\)6.4&66.3\(\pm\)4.9&52.3\(\pm\)3.3&64.0\(\pm\)9.8&\textbf{67.3\(\pm\)5.0}&59.0\(\pm\)15.7\\
			block hammer beat&59.0\(\pm\)5.3&66.3\(\pm\)1.3&62.0\(\pm\)9.2&74.0\(\pm\)7.8&63.3\(\pm\)3.4&\textbf{81.3\(\pm\)9}\\
			dual bottles pick (easy)&90.0\(\pm\)3.5&\textbf{91.7\(\pm\)0.9}&86.0\(\pm\)2.9&89.3\(\pm\)2.5&89.7\(\pm\)4.1&89.3\(\pm\)0.5\\
			diverse bottles pick&45.7\(\pm\)12.0&55.3\(\pm\)5.3&53.7\(\pm\)4.0&\textbf{58.3\(\pm\)3.3}&52.3\(\pm\)4.9&55.0\(\pm\)1.4\\
			pick apple messy &7.7\(\pm\)3.4&\textbf{14.3\(\pm\)8.0}&9.0\(\pm\)4.2&11.0\(\pm\)6.5&6.3\(\pm\)1.7&13.7\(\pm\)9.2\\
			\hline
			average&48.5& 58.8 &52.6 &59.2&55.8&\textbf{59.7}\\
			\hline
		\end{tabular}
	\end{center}
\end{table*}
% 为了确定所提出的自适应梯度调整方法是否能有效提高算法的性能，我们在Robotwin上基于五个任务进行了消融实验。详细的实验结果见表~\ref{tabaga}。我们发现，在去除梯度调整后，一些任务的性能急剧下降。比如在block hammer beat任务上，性能下降达到了65.2\%,在其他四个任务上也有不同程度的降低。实验结果表明，本文提出的自适应梯度调整算法能够很好地综合这两种损失，提高算法性能。

To determine whether the proposed adaptive gradient adjustment method can effectively improve the algorithm's performance, we conducted ablation experiments based on five tasks on Robotwin. The detailed experimental results are shown in Table~\ref{tabaga}. We found that after removing the gradient adjustment, the performance on some tasks dropped sharply. For example, on the block hammer beat task, the performance decline reached 65.2\%, and there were also different degrees of reduction on the other four tasks. These experimental results indicate that the adaptive gradient adjustment algorithm proposed in this paper can well integrate the two losses and improve the algorithm's performance.

\begin{table}
	\begin{center}
		\caption{The influence of AGA}
		\label{tabaga}
		\setlength{\tabcolsep}{3pt}
		\begin{tabular}{p{85pt}p{42pt}p{42pt}}
			\hline
			Task&with & without\\
			\hline
			bottle adjust&59.0\(\pm\)15.7&\textbf{60.6\(\pm\)5.4}\\
			block hammer beat&\textbf{81.3\(\pm\)0.9}&28.3\(\pm\)17.8\\
			dual bottles pick (easy)&\textbf{89.3\(\pm\)0.5}&84.6\(\pm\)6.9\\
			diverse bottles pick&\textbf{55.0\(\pm\)1.4}&48.6\(\pm\)4.0\\
			pick apple messy&\textbf{13.7\(\pm\)9.2}&10.3\(\pm\)4.8\\
			\hline
		\end{tabular}
	\end{center}
\end{table}

% 为了跟进一步的确定所提出的方法有效性的原因，我们对神经网络的land space进行了可视化。具体来说，我们在两个任务上采用land space提出的方法，以单步预测的动作误差作为度量指标进行看可视化，其结果如图所示。 我们对比使用和不使用ADA算法所得到的error landscape，从3D形貌（i)）来看，加入之后，误差的3D形貌中没有平原区域，并且误差也更小，从等高线图（ii)和iii)）来看，加入ADA算法之后，等高线更加稀疏，这也就意味着变化更加平滑，并且同等误差覆盖范围更广。这些对比表明，我们所提出的ADA算法增强了网络的泛化性。
\begin{figure}[htpb]
	\centering
	\subfloat[ With AGA]{{\includegraphics[width=0.9\columnwidth]{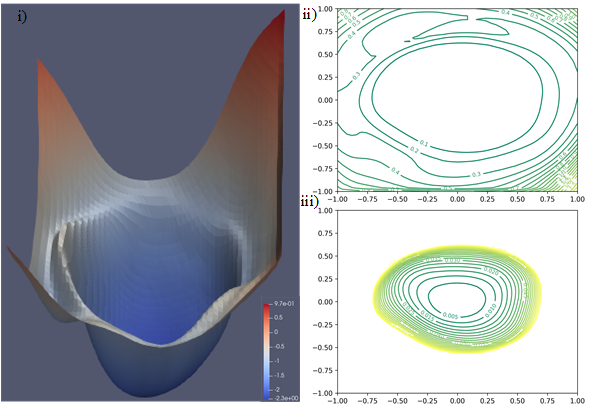}}} \vfil
	\subfloat[Without AGA]{{\includegraphics[width=0.9\columnwidth]{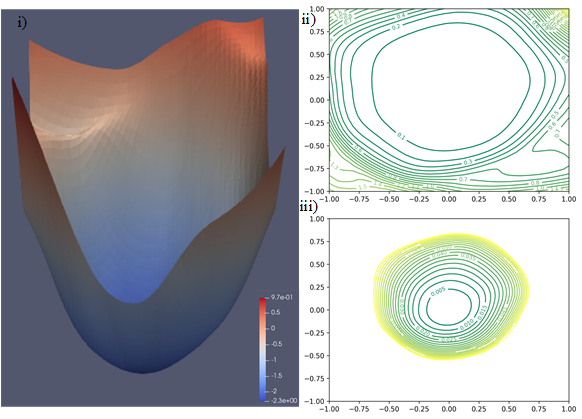}}}
	\caption{The Landscape of Neural Nets in bloack hammer.}
	\label{visland}
\end{figure}

%To further investigate the underlying mechanisms behind the effectiveness of our proposed method, we performed visualization of the neural network's error landscape. Specifically, we applied the proposed technique in~\cite{NEURIPS2018a41b3bb3} to the block hammer task and conducted visualizations using single-step prediction error as the metric, with results illustrated in the Figure~\ref{visland}. We conducted comparative analyses of error landscapes with and without the AGA algorithm. From the 3D topography perspective i), the incorporation of AGA eliminates plateau regions in the error surface while achieving lower overall error values. Observations from the contour maps ii) and iii) further demonstrate sparser contour distributions after applying AGA, indicating smoother transitions and broader coverage under equivalent error thresholds. These comparative results substantiate that our proposed AGA algorithm enhances the network's generalization capacity.

To further explore the mechanisms underlying the effectiveness of our method, we visualized the neural network’s error landscape. Specifically, we applied the technique proposed in~\cite{NEURIPS2018a41b3bb3} to the block-hammer task, visualizing one-step prediction error as the metric; results are shown in Fig.~\ref{visland}. We compared the error landscapes with and without the AGA algorithm. From the 3D topography (i), incorporating AGA eliminates plateau regions in the error surface, yielding lower overall error values. The contour maps (ii) and (iii) further reveal that applying AGA results in sparser contours, indicating smoother transitions and broader coverage under equivalent error thresholds. These results confirm that the proposed AGA algorithm reduces the one-step error and improves the network’s generalization capability

%

%To understand the influence of the initial value of \(c\) in Eq~\ref{c_eq} on the algorithm's performance, we also conducted corresponding ablation experiments. As before, we selected five tasks on Robotwin to carry out relevant ablation experiments, and the results are shown in Table~\ref{tabc} . We found that the initial value has a limited impact on most tasks, but has a relatively large impact on the block hammer beat task. Combined with the previous experimental results, this suggests that the task is more challenging to optimize and is highly sensitive to the gradient.

To examine how the initial value of \(c\) in Eq.~\ref{c_eq} affects algorithm performance, we conducted additional ablation experiments. As before, five Robotwin tasks were used for ablation, and the results are presented in Table~\ref{tabc}. The initial value shows minimal impact on most tasks but a noticeable effect on the block-hammer-beat task. Combined with previous results, this indicates that the task is more difficult to optimize and highly sensitive to gradient variations.

\begin{table}
	\begin{center}
		\caption{The influence of the initial threshold.}
		\label{tabc}
		\setlength{\tabcolsep}{3pt}
		\begin{tabular}{p{85pt}p{42pt}p{42pt}p{42pt}}
			\hline
			Task&1.0&0.5&0.01\\
			\hline
			bottle adjust&59.0\(\pm\)15.7&\textbf{63.7\(\pm\)5.9} &61.0\(\pm\)7.3\\
			block hammer beat&\textbf{81.3\(\pm\)9}&73.7\(\pm\)8.8&64.6\(\pm\)6.5\\
			dual bottles pick (easy)&\textbf{89.3\(\pm\)0.5}&88.0\(\pm\)2.9&85.7\(\pm\)7.6\\
			diverse bottles pick&\textbf{55.0\(\pm\)1.4}&53.3\(\pm\)1.9&53.6\(\pm\)2.6\\
			pick apple messy&13.7\(\pm\)9.2&\textbf{15.0\(\pm\)7.5}&14.3\(\pm\)5.7\\
			\hline
		\end{tabular}
	\end{center}
\end{table}

%To determine the influence of NFE on the final performance of the algorithm during the inference stage, we conducted relevant ablation studies on five Robotwin tasks. We found that the algorithm proposed in this paper can achieve high-performance inference with a relatively small number of steps. As the number of steps increases, the performance of the algorithm first increases and then decreases.

% 为了探究NFE在推理阶段对算法最终性能的影响，我们针对5个Robotwin任务进行了相关的消融研究。实验结果如表~\ref{step}所示，我们发现，本文提出的算法能够以相对较少的步数实现高性能的推理。值得注意的是，不同的任务上算法的表现也不一样，比如在bottle adjust任务上（这是我们与3DP差距最大的任务），随着推理步数的增加，算法的性能也在增加，最终在10步的时候也能够达到和3DP一样的性能。在diverse botles pick任务中，算法的性能却在下降，但是下降幅度并不大。总的来说，算法的性能随着推理步数增加呈现出先增后降的趋势。

%To examine how the number of function evaluations (NFE) influences the algorithm’s inference performance, ablation studies were conducted on five Robotwin tasks. As shown in Table~\ref{step}, the proposed algorithm achieves strong inference performance with relatively few steps. Notably, the algorithm’s performance varies across tasks. For example, in the bottle adjustment task—where the largest gap relative to 3DP is observed—performance improves as the number of inference steps increases, eventually matching 3DP at ten steps. Conversely, in the diverse bottle-picking task, performance slightly decreases, though the decline remains modest.
To examine how the NFE affects the algorithm’s inference performance, we conducted ablation studies on five Robotwin tasks.  As shown in Table~\ref{step}, the proposed algorithm achieves strong inference performance with relatively few steps. Notably, the algorithm’s performance varies across tasks. For example, in the bottle adjustment task—where the largest gap relative to 3DP is observed—performance improves as the number of inference steps increases, eventually matching 3DP at ten steps. Conversely, in the diverse bottle picking task, performance slightly decreases, though the decline remains modest. Overall, the algorithm exhibits a trend: performance improves initially but gradually declines as inference steps increase.

\begin{table}[htpb]
	\begin{center}
		\caption{The influence of the inference step.}
		\label{step}
		\setlength{\tabcolsep}{3pt}
		\begin{tabular}{p{85pt}p{35pt}p{35pt}p{35pt}p{35pt}}
			\hline
			Task&1&3&5&10\\
			\hline
			bottle adjust&59.0\(\pm\)15.7&64.0\(\pm\)11.3 &67.7\(\pm\)11.5&\textbf{71.7\(\pm\)10.2}\\
			block hammer beat&81.3\(\pm\)9&80.0\(\pm\)5.0&\textbf{82.3\(\pm\)1.7}&77.3\(\pm\)2.6\\
			dual bottles pick (easy)&89.3\(\pm\)0.5&91.7\(\pm\)2.5&91.0\(\pm\)2.4&\textbf{92.0\(\pm\)2.2}\\
			diverse bottles pick&\textbf{55.0\(\pm\)1.4}&51.7\(\pm\)3.6&54.0\(\pm\)2.8&51.3\(\pm\)0.4\\
			pick apple messy&\textbf{13.7\(\pm\)9.2}&12.3\(\pm\)8.7&12.0\(\pm\)8.6&12.3\(\pm\)8.0\\
			\hline
			average& 59.7&59.9&\textbf{61.4}&60.9\\
			\hline
		\end{tabular}
	\end{center}
\end{table}

%Our experiments have comprehensively answered the four questions in this section. Regarding the first question, the experimental results show that the proposed method can handle various manipulation tasks. For the second question, we have demonstrated the effectiveness of the proposed method in the real-world through three experiments. As for the third question, our experiments on two rich-contact manipulation tasks in the real-world indicate that the proposed method can be well generalized to multimodal fusion perception tasks. Concerning the last question, we have verified through ablation experiments that the treatment of gradients can indeed improve the performance of the algorithm.

The experiments comprehensively address the four research questions presented in this section. For the first question, results demonstrate that the proposed method effectively handles diverse manipulation tasks. For the second question, three real-world experiments confirm the effectiveness of the proposed method. For the third question, real-world experiments on two rich-contact manipulation tasks indicate that the method generalizes well to multimodal fusion perception. Finally, ablation experiments verify that gradient treatment indeed enhances algorithm performance.

\section{Conclusion}

%In this paper, we propose a one-step flow matching algorithm based on multi-step integration and successfully apply it to robot tasks, achieving competitive results. To enhance the stability of the training process, we propose an adaptive gradient allocation method, regarding the flow matching loss and the consistency loss as a special type of multi-task learning. In addition, we verify the effectiveness of the proposed method through a large number of simulation and real-world experiments. Especially, the experimental results on the rich contact tasks integrating visual-tactile perception show that the proposed method has strong competitiveness in such tasks. Finally, we conduct ablation experiments on each component of the proposed method to verify the performance of each part. In the future, we will combine vision-language large models to conduct relevant research to improve the inference speed of related tasks.

This study proposes a one-step flow-matching algorithm built upon multi-step integration and successfully applies it to robotic tasks, achieving competitive performance.  To enhance training stability, an adaptive gradient allocation method is introduced, which treats the flow-matching and consistency losses as a specific form of multi-task learning. The effectiveness of the proposed method is further validated through extensive simulation and real-world experiments. In particular, results from rich-contact tasks integrating visual–tactile perception demonstrate the strong competitiveness of the proposed method. Finally, ablation experiments are conducted on each component to evaluate its individual contribution to overall performance. Future work will explore integrating vision–language large models to further enhance inference efficiency in related tasks.

%\section*{Acknowledgments}
%This should be a simple paragraph before the References to thank those individuals and institutions who have supported your work on this article.

%{\appendix[A]
%
%}

{\appendices
\section*{Appendix A}
\label{appendixa}
Here we give the detailed derivation of the closed-form solution of our AGA. The goal of our method is:
\begin{equation}
	c(\mathbf{g}\mathbf{u}^{\top}_1)=\mathbf{g}\mathbf{u}^{\top}_2
\end{equation}

Given conditions: \( \mathbf{g} = \alpha_1 g_1 + \alpha_2 g_2 \), \( \alpha_1 + \alpha_2 = 1 \), \( g_1 = A u_1 \), \( g_2 = B u_2 \), \( \delta = u_1 \cdot u_2 \). Substituting \( \mathbf{g} = \alpha_1 g_1 + \alpha_2 g_2 \) and \( \delta = u_1 \cdot u_2 \), we obtain:

\begin{equation}
	c \cdot [\alpha_1 A + \alpha_2 B \delta] = \alpha_1 A \delta + \alpha_2 B
\end{equation}

% 代入
Substituting \( \alpha_2 = 1 - \alpha_1 \) yields:
\begin{equation}
	c (\alpha_1 A + (1 - \alpha_1) B \delta) = \alpha_1 A \delta + (1 - \alpha_1) B
\end{equation}
% 然后开展化简可以得到：
Then, after simplification, we get:
\begin{equation}
	\alpha_1 (c A - c B \delta) + c B \delta = \alpha_1 (A \delta - B) + B
\end{equation}
% 最终得到
Finally, we can get
\begin{equation}
	\alpha_1 = \frac{B (1 - c \delta)}{(c A - A \delta) + B(1 - c \delta)}
\end{equation}
\section*{Appendix B}
\label{appendixb}
The constraint that the value of \( c \) should satisfy during the dynamic adjustment process can be derived based on \( 0 < \alpha_1 < 1 \) and \( \alpha_1 = \frac{B (1 - c \delta)}{(c A - A \delta) + B(1 - c \delta)} \). Thus, we obtain
\begin{equation}
	0<\frac{B (1 - c \delta)}{A(c  -  \delta) + B(1 - c \delta)}<1
\end{equation}

% 因为我们multi-consist loss是依赖于flow-macthing loss的，因此我们定义c处于[0,01,1]之间，始终保证flow-macthing loss主导合成梯度的方向。根据c的定义区间，分子天然为正。根据\(\frac{B (1 - c \delta)}{(c A - A \delta) + B(1 - c \delta)}<1\)，我们可以直接得出第一个约束：

Since the multi-step consistency loss depends on the flow-matching loss, we define \( c \) to be within the interval \([0,1]\), always ensuring that the flow-matching loss dominates the direction of the composite gradient. Given the definition interval of \( c \), the numerator is naturally positive. Based on \(\frac{B (1 - c \delta)}{(c A - A \delta) + B(1 - c \delta)} < 1\), we can directly derive the first constraint:
\begin{equation}
	\delta<c
\end{equation}
% 我们先考虑\(\delta > 0\)的情况
% 由于\(B (1 - c \delta)>0\)，因此：
Given that \( B (1 - c \delta) > 0 \), it follows that:
\begin{equation}
	c(A-B\delta)>(A\delta-B)
\end{equation}
% 当\(A>B\delta\)时，我们有：
When \( A > B\delta \), we have
\begin{equation}
	c>\frac{(A\delta-B)}{(A-B\delta)}
\end{equation}
% 当\(A<B\delta\)时，我们有：
When \( A < B\delta \), we have
\begin{equation}
	c<\frac{(A\delta-B)}{(A-B\delta)}
\end{equation}
% 当\(A=B\delta)时，由于\(-1<\delta<1\)，于是这种情况下\(0<\delta<1，A<B\)，因此\(A\delta-B<0\)成立，此时有：

When \( A = B\delta \), given that \( -1 < \delta < 1 \), in this case \( 0 < \delta < 1 \) and \( A < B \), thus \( A\delta - B < 0 \) holds. At this case, we have:
\begin{equation}
	\frac{B (1 - c \delta)}{B-A\delta}<1
\end{equation}
% 可以得到
Then we have :
\begin{equation}
	\frac{A}{B}<c
\end{equation}
% 当\(\delta<=0\)时，\(c>0\)即可。
% When \(\delta \leq 0\), \( c > 0 \) is sufficient.

% 综上，我们可以得到c的所有约束条件：
In summary, we can derive all the constraints for \( c \):

\begin{equation}
	\left\{
	\begin{aligned}
		& c <  \frac{A\delta-B}{A-B\delta} \quad &\textbf{if} \quad \delta < c \quad \textbf{and} A<B\delta\\
		&\frac{A}{B} \quad < c \quad &\textbf{if} \quad \delta < c \quad \textbf{and} A=B\delta\\
		& \frac{A\delta-B}{A-B\delta} \quad < c \quad &\textbf{if} \quad \delta < c \quad \textbf{and} A>B\delta\\
	\end{aligned}
	\right.
\end{equation}}

\bibliography{./BIBTCyber}

\end{document}